\documentclass[eat,twocolumn]{jmlr}

\usepackage{longtable}

\usepackage{booktabs}
\usepackage[load-configurations=version-1]{siunitx} 

\theorembodyfont{\upshape}
\theoremheaderfont{\scshape}
\theorempostheader{:}
\theoremsep{\newline}

\jmlrvolume{}
\firstpageno{1}

\jmlryear{2022}
\jmlrworkshop{Machine Learning for Health (ML4H) 2022}

\title[Machine Learning Localizes Seizure Onset Zone]{Machine Learning Methods Applied to Cortico-Cortical Evoked Potentials Aid in Localizing Seizure Onset Zones}

\author{\Name{Ian G. Malone} \Email{ian.g.malone@gmail.com} \textnormal{\emph{University of Florida}}\\
\Name{Kaleb E. Smith} \Email{kasmith@nvidia.com} \textnormal{\emph{NVIDIA}}\\
\Name{Morgan E. Urdaneta} \Email{morganu7@gmail.com} \textnormal{\emph{University of Florida}}\\
\Name{Tyler S. Davis} \Email{tyler.davis@hsc.utah.edu} \textnormal{\emph{University of Utah}}\\
\Name{Daria Nesterovich Anderson} \Email{daria.anderson@hsc.utah.edu} \textnormal{\emph{University of Utah}}\\
\Name{Brian J. Phillip} \Email{brian.philip@utah.edu} \textnormal{\emph{University of Utah}}\\
\Name{John D. Rolston} \Email{jrolston@bwh.harvard.edu} \textnormal{\emph{Harvard University}}\\
\Name{Christopher R. Butson} \Email{butsonc@ufl.edu} \textnormal{\emph{University of Florida}}\\
}

\begin{document}
\raggedbottom
\maketitle

\begin{abstract}
Epilepsy affects millions of people, reducing quality of life and increasing risk of premature death. One-third of epilepsy cases are drug-resistant and require surgery for treatment, which necessitates localizing the seizure onset zone (SOZ) in the brain. Attempts have been made to use cortico-cortical evoked potentials (CCEPs) to improve SOZ localization but none have been successful enough for clinical adoption. Here, we compare the performance of ten machine learning classifiers in localizing SOZ from CCEP data. This preliminary study validates a novel application of machine learning, and the results establish our approach as a promising line of research that warrants further investigation. This work also serves to facilitate discussion and collaboration with fellow machine learning and/or epilepsy researchers.
\end{abstract}


\begin{figure*}[htbp]
\centering
  {\caption{CCEPs are recorded from many electrodes in the brain. Metadata are added to the dataset along with a binary target variable. Data are cleaned, oversampled, and encoded. Seven different train-test splits are fed into ten classifiers for performance comparison.}}
  {\includegraphics[width=.90\linewidth]{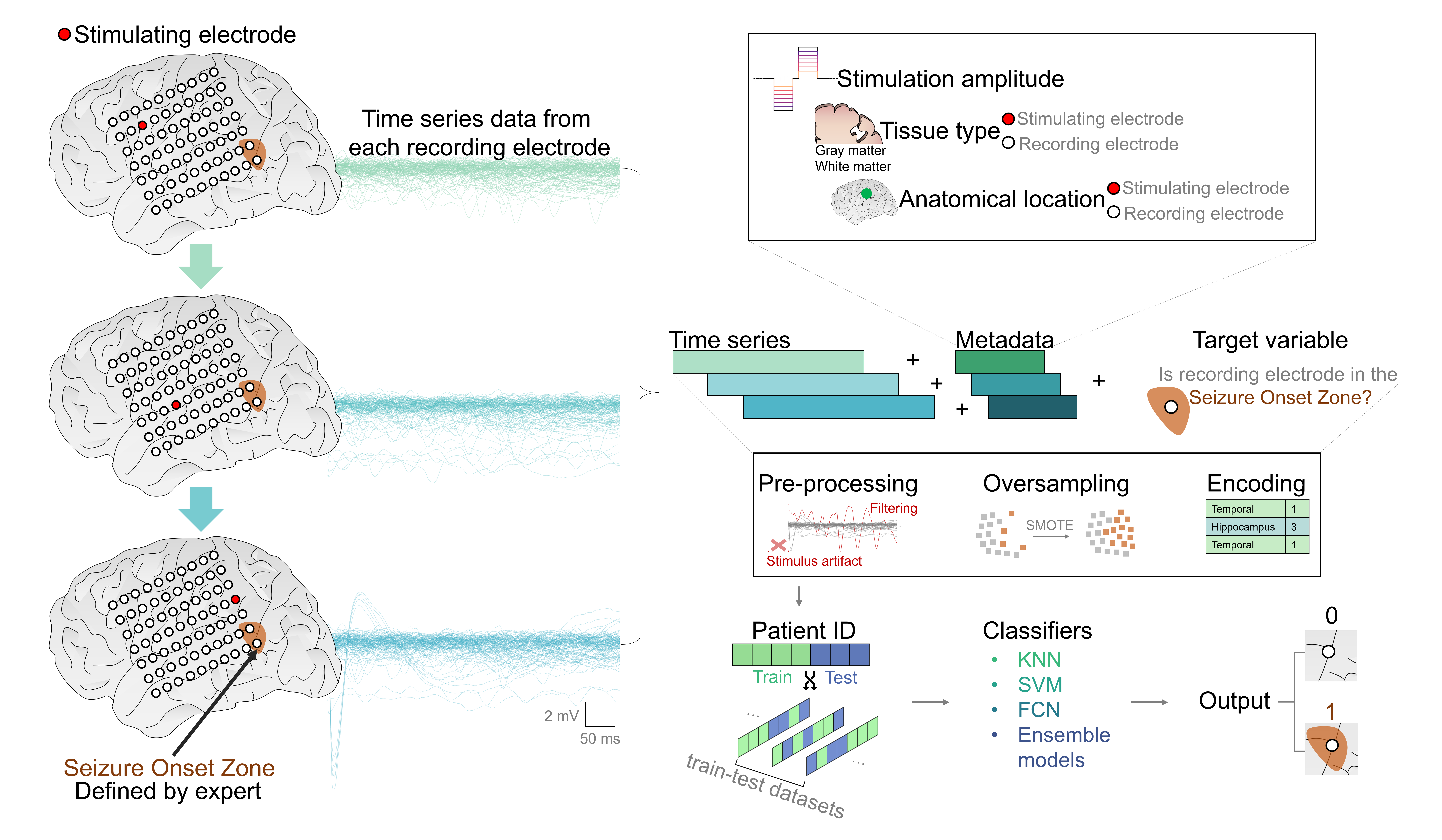}}
  
\end{figure*}

\section{Introduction}
Epilepsy—one of the most common neurological diseases, globally—affects 1 in 26 people during their lifetimes, causing an up to 3 times greater risk of premature death than in the general population \citep{who}. One in three people with epilepsy continue to have seizures despite medication. For these patients with drug-resistant epilepsy, surgery is their only hope for a cure. However, epilepsy surgery is only effective if the seizure onset zone (SOZ)—where a seizure originates in the brain—is precisely localized. To localize the SOZ, brain activity must be properly recorded and analyzed. Patients must stay in the clinic for several days while connected to recording hardware via electrodes that penetrate the skull and brain. This method, which has been used for decades, may benefit from leveraging modern computational techniques to boost localization efficiency. Improvements here could profoundly augment patient quality of life via reduced localization time, cost, and risk.

One method of probing the brain's dynamic function is to apply single pulses of electrical stimulation and record the ensuing brain activity. These recordings are termed cortico-cortical evoked potentials (CCEPs) \citep{keller}. Various CCEP parameters have been used to help localize the SOZ, such as amplitude \citep{kundu}, spectral power \citep{davis}, time-frequency \citep{xiao}, and graph theoretic measures \citep{parker}, though none have been successful enough to be used in clinical practice. Using the entire CCEP time series may have advantages over using a reduced set of CCEP features, as more information will be preserved in the input. However, this approach vastly increases the dimensionality of the feature space, which can make traditional statistical inference prohibitively challenging. Machine learning may be particularly well suited to handle this challenge, as many models have the capability of learning nonlinear input-output mappings in large datasets and exploiting information that would otherwise go unused with more human-directed approaches.

Here, we compare the performance of ten machine learning classifiers in localizing the SOZ in our CCEP dataset. The main purpose of this work is to serve as a preliminary study to establish precedent for applying machine learning in this problem domain. Strong preliminary results here will justify a follow-up study from our group that leverages more advanced modeling and a dataset with larger cohorts and enriched metadata gleaned from functional neuroimaging studies. 

\section{Data Collection}
The data used here were collected from 2019–2020 (approved by the Institutional Review Board; IRB protocol 00069440). Data were recorded from seven patients during intracranial monitoring for medically refractory epilepsy while patients were awake and resting. Stereotactic electroencephalography electrodes implanted for routine medical management were used for both stimulation and recording. Each patient had 50--90 electrodes implanted. The SOZ was determined by the treating epileptologist, who labeled each electrode as being in- or outside of the SOZ after reviewing recordings of seizure activity. This labelling proceudre is standard practice in the clinic. To record CCEPs, a single cathode-leading biphasic pulse (0.5 ms/phase) with a randomized interpulse interval (2.5–3.5 s) was applied to a single electrode, and all other electrodes recorded the ensuing neurophysiological activity. This procedure was repeated such that each electrode in a single patient operated as the stimulation electrode once. This dataset also contains several metadata features that give context to each recording, including: the stimulation amplitude; the anatomical location of the electrodes (Brainnetome Atlas \citep{brain}); and the tissue type (gray or white matter) and side of the brain (left or right hemisphere) that the electrodes are in. The dataset is 653,764 rows by 503 columns. There are 495 columns—corresponding to 495 ms—of time series data, 7 columns of metadata, and the single binary target column indicating whether an electrode is in the SOZ. Each row corresponds to a unique CCEP. For additional information on electrode specifications, locations, and stimulation and recording paradigms, please see \citep{kundu}. Code for this project is available at \href{https://github.com/IanGMalone/ML4Hsoz}{github.com/IanGMalone/ML4Hsoz}. 

\begin{table*}[htbp]
\centering
\caption{Average performance metrics over seven splits of data for each model. The three highest values of each metric are bolded.}
\label{tab:table1}
\begin{tabular}{@{}ccccc@{}}
\toprule
\textbf{Model}      & \textbf{Macro Precision} & \textbf{Macro Recall} & \textbf{ROC AUC}   & \textbf{Accuracy}   \\ \midrule
KNN                 & 60.1 $\pm$1.96          & 64.9 $\pm$2.26           & 71.1 $\pm$2.62          & 85.6 $\pm$ 1.83            \\
FCN-TS              & 52.8 $\pm$ 1.38         & 56.6 $\pm$1.77           & 77.3 $\pm$2.02          & 69.3 $\pm$1.85               \\
FCN-TSM             & \textbf{69.4 $\pm$1.67} & 67.1 $\pm$1.86           & \textbf{80.8 $\pm$1.95} & \textbf{88.9 $\pm$1.90}       \\
SVM-Poly            & 60.2 $\pm$2.07          & 63.3 $\pm$2.17           & 69.4 $\pm$2.75          & 87.7 $\pm$2.23               \\
SVM-Rbf             & 59.3 $\pm$2.43          & 64.3 $\pm$2.32           & 70.1 $\pm$2.57          & 85.4 $\pm$2.44               \\
Random Forest       & 65.6 $\pm$2.11     & {\textbf{76.6 $\pm$1.99}}     & 74.5 $\pm$2.09          & 86.1 $\pm$2.01              \\
Extra Tree Boosting & 64.6 $\pm$2.09     & \textbf{74.9 $\pm$2.02}       & 78.1 $\pm$2.15          & 85.4 $\pm$1.99               \\
XGBoost             & {\textbf{74.4 $\pm$1.77}} & 58.4 $\pm$1.93         & 79.1 $\pm$2.23     &{\textbf{90.7$\pm$1.89}} \\
CatBoost            & 66.1 $\pm$1.80          & 67.0 $\pm$1.90           & \textbf{81.7 $\pm$2.11} & \textbf{89.0$\pm$2.07}       \\
Soft Ensemble       & \textbf{68.2 $\pm$1.56} & \textbf{76.3 $\pm$1.83}  & {\textbf{83.2 $\pm$1.99}} & 87.7 ~1.79             \\ \bottomrule
\end{tabular}
\end{table*}

The first 5 ms of each recording was removed to account for stimulation artifact in the CCEP. Recordings were then visually inspected. Those having clearly non-biological characteristics (e.g., saturated amplifier and other obvious artifacts) were removed from the dataset. Categorical features were encoded using target encoding to preserve the dimensionality of the original dataset. To better understand how epilepsy’s characteristic heterogeneity \citep{pennell} impacts model performance, we created seven distinct train and test datasets. In each of the seven datasets, four of the seven patients were randomly allocated to the training set, and the remaining three to the testing set, to prevent data leakage. The dataset was highly imbalanced, with roughly eight percent of the electrodes being in the SOZ (class 1). Thus, we over-sampled the training set’s minority class using the Synthetic Minority Over-sampling Technique \citep{chawla} to achieve an equal distribution of positive and negative class records. 

\section{Experimental Setup}
We compared ten models with minimal tuning (Table 1): eight machine learning models; a deep learning model; and an ensemble model. The machine learning models were: K-Nearest Neighbors (K=3) with Dynamic Time Warping; Extra Tree Boosting (500 estimators); Random Forest (500 estimators); XGBoost (1200 estimators) \citep{xgboost}; SVM with 5-degree polynomial kernel; SVM with Radial Basis Function kernel; and CatBoost (1000 estimators) \citep{catboost}. Our deep learning model was a simple three-layer 1D-CNN with global average pooling, termed a Fully Connected Network (FCN) \citep{wang}. Each layer had 64 filters of sizes 7, 5, and 3, respectively. Our ensemble model combined Extreme Tree Boosting, Random Forest, XGboost, and CatBoost results in a soft voting manner (Soft Ensemble). We chose this set of popular models to broadly gauge if this SOZ localization problem is amenable to a machine learning approach. To test whether metadata improves predictive power, we compared the performance of the FCN trained only on the time series data (FCN-TS) to the FCN trained on the time series with concatenated metadata (FCN-TSM) (Table 1). The FCN trained with time series and metadata performed better; thus, all other models were trained in the same manner. Precision, recall, and F1-score are reported in addition to accuracy. Accuracy here may be misleading due to the aforementioned class imbalance. Our experiments were run on the University of Florida's DGX-A100 SuperPod. Using NVIDIA’s RAPIDS AI \citep{rapids-ai} and Tensorflow containers \citep{NGC}, algorithms ran on a single A100 GPU with 80 GB of VRAM.

\section{Results and Discussion}
Table 1 shows performance metrics for each model averaged across the seven different train-test splits. A simple KNN was used as a benchmark. Two notable results appear here. First, including both metadata and time series when training the FCN (FCN-TSM) outperforms the FCN-TS, which was trained solely on time series. This establishes the importance of giving context to the time series with metadata and supports future endeavors to further enrich the metadata. Second, the Soft Ensemble model has the highest performance across the most metrics (i.e., three metrics with performance scores in the top 3), and these results are comparable to those of other published methods \citep{johnson}. Though this performance is far from ideal, it could provide utility in a clinical setting as a measure of confidence to augment a clinician's judgment. Nonetheless, much more work is needed to develop and validate this approach before such an implementation. We will further investigate ensemble models in our coming study as they may generalize well despite the heterogeneity inherent in epilepsy data.

These exciting preliminary results establish precedent for applying machine learning for SOZ localization, justifying forthcoming efforts from our group to incorporate larger patient cohorts. We hypothesize that this larger dataset will markedly improve model performance due to epilepsy’s known inter-patient variability \citep{pennell}. This larger dataset also has drastically enriched metadata imported from neuroimaging studies which quantifies distance and probability of functional connection between each recording and stimulating electrode pair. Performance could also be increased via augmenting ground truth SOZ labels with more epileptologists' opinions.

We are encouraged by the respectable performance of these off-the-shelf classifiers and will invest effort in more advanced modeling that accounts for temporal dependence in the data rather than the current approach, which treats each time sample as a unique feature. We are also developing a novel deep learning approach that determines if a batch of recordings belongs to a single electrode in the SOZ. We will also further examine ensemble models and test the performance of models ingesting a simple tabular dataset consisting of metadata and descriptive features extracted from the time series. An additional question to investigate is the utility of biasing the model toward precision or recall based on epilepsy-specific factors.

The impact of this work is threefold: 1) Novelty: This work applies machine learning in a novel way to CCEP time series and metadata for SOZ localization; 2) Future work: These exciting preliminary results justify increased efforts in improving our modeling and applying it to larger datasets; 3) Collaboration - This work opens the door for discussion and collaboration with researchers in this space. With this promising line of research, we aim to augment clinicians’ capabilities to efficiently localize SOZ by reducing the associated time, cost, and risk while simultaneously improving surgical outcomes. Success here could improve life for millions of people currently suffering from epilepsy.

\label{sec:cite}





\bibliography{pmlr-sample}

\begin{thebibliography}{15}
\providecommand{\natexlab}[1]{#1}
\providecommand{\url}[1]{\texttt{#1}}
\expandafter\ifx\csname urlstyle\endcsname\relax
  \providecommand{\doi}[1]{doi: #1}\else
  \providecommand{\doi}{doi: \begingroup \urlstyle{rm}\Url}\fi

\bibitem[Chawla et~al.(2002)Chawla, Bowyer, Hall, and Kegelmeyer]{chawla}
N.~V. Chawla, K.~W. Bowyer, L.~O. Hall, and W.~P. Kegelmeyer.
\newblock Smote: Synthetic minority over-sampling technique.
\newblock \emph{J. AI Research}, 16:\penalty0 321--357, January 2002.
\newblock URL \url{https://doi.org/10.1613%2Fjair.953}.

\bibitem[Chen and Guestrin(2016)]{xgboost}
T.~Chen and C.~Guestrin.
\newblock {XGBoost}: A scalable tree boosting system.
\newblock In \emph{Proceedings of the 22nd ACM SIGKDD International Conference
  on Knowledge Discovery and Data Mining}, KDD '16, pages 785--794, New York,
  NY, USA, 2016. ACM.
\newblock ISBN 978-1-4503-4232-2.
\newblock \doi{10.1145/2939672.2939785}.
\newblock URL \url{http://doi.acm.org/10.1145/2939672.2939785}.

\bibitem[Davis et~al.(2018)Davis, Rolston, Bollo, and House]{davis}
T.~S. Davis, J.~D. Rolston, R.~J. Bollo, and P.~A. House.
\newblock Delayed high-frequency suppression after automated single-pulse
  electrical stimulation identifies the seizure onset zone in patients with
  refractory epilepsy.
\newblock \emph{Clin Neurophysiol.}, 129:\penalty0 2466--2474, November 2018.
\newblock URL \url{https://doi.org/10.1016/j.clinph.2018.06.021}.

\bibitem[Dorogush et~al.(2017)Dorogush, Gulin, Gusev, Kazeev, Prokhorenkova,
  and Vorobev]{catboost}
A.~V. Dorogush, A.~Gulin, G.~Gusev, N.~Kazeev, L.~O. Prokhorenkova, and
  A.~Vorobev.
\newblock Fighting biases with dynamic boosting.
\newblock \emph{CoRR}, abs/1706.09516, 2017.
\newblock URL \url{http://arxiv.org/abs/1706.09516}.

\bibitem[Fan et~al.(2016)Fan, Li, Zhuo, Zhang, Wang, Chen, Yang, Chu, Xie,
  Laird, Fox, Eickhoff, Yu, and Jiang]{brain}
L.~Fan, H.~Li, J.~Zhuo, Y.~Zhang, J.~Wang, L.~Chen, Z.~Yang, C.~Chu, S.~Xie,
  A.~R. Laird, P.~T. Fox, S.~B. Eickhoff, C.~Yu, and T.~Jiang.
\newblock The human brainnetome atlas: A new brain atlas based on connectional
  architecture.
\newblock \emph{Cereb Cortex.}, August 2016.
\newblock URL \url{https://doi.org/10.1093/cercor/bhw157}.

\bibitem[Johnson et~al.(2022)Johnson, Cai, Doss, Jiang, Negi, Narasimhan,
  Paulo, Gonzalez, Roberson, Bick, Chang, Morgan, Wallace, and Englot]{johnson}
G.~W. Johnson, L.~Y. Cai, D.~J. Doss, J.~W. Jiang, A.~S. Negi, S.~Narasimhan,
  D.~L. Paulo, H.~F.~J. Gonzalez, S.~W. Roberson, S.~K. Bick, C.~E. Chang,
  V.~L. Morgan, M.~T. Wallace, and D.~J. Englot.
\newblock Localizing seizure onset zones in surgical epilepsy with
  neurostimulation deep learning.
\newblock \emph{J. Neurosurg}, pages 1--6, 2022.
\newblock URL \url{https://doi.org/10.3171/2022.8.JNS221321}.

\bibitem[Keller et~al.(2015)Keller, Honey, Mégevand, Entz, Ulbert, and
  Mehta]{keller}
C.~J. Keller, C.~J. Honey, P.~Mégevand, L.~Entz, I.~Ulbert, and A.~D. Mehta.
\newblock Mapping human brain networks with cortico-cortical evoked potentials.
\newblock \emph{Philos Trans R Soc Lond B Biol Sci.}, 369, October 2015.
\newblock URL \url{https://doi.org/10.1098/rstb.2013.0528}.

\bibitem[Kundu et~al.(2020)Kundu, T.~S. Davis~and, Smith, Arain, Peters,
  Newman, Butson, and Rolston]{kundu}
B.~Kundu, B.~Philip T.~S. Davis~and, E.~H. Smith, A.~Arain, A.~Peters,
  B.~Newman, C.~R. Butson, and J.~D. Rolston.
\newblock A systematic exploration of parameters affecting evoked intracranial
  potentials in patients with epilepsy.
\newblock \emph{Brain Stimul.}, 13:\penalty0 1232--1244, June 2020.
\newblock URL \url{https://doi.org/10.1016/j.brs.2020.06.002}.

\bibitem[NVIDIA(2022)]{NGC}
NVIDIA.
\newblock Nvidia ngc | catalog.
\newblock Technical report, NVIDIA, 2022.
\newblock URL
  \url{https://catalog.ngc.nvidia.com/orgs/nvidia/containers/tensorflow}.

\bibitem[Parker et~al.(2017)Parker, Clayden, Cardoso, Rodionov, Duncan, Scott,
  Diehl, and Ourselin]{parker}
C.~S. Parker, J.~D. Clayden, M.~J. Cardoso, R.~Rodionov, J.~S. Duncan,
  C.~Scott, B.~Diehl, and S.~Ourselin.
\newblock Structural and effective connectivity in focal epilepsy.
\newblock \emph{Neuroimage Clin.}, 17:\penalty0 943--952, December 2017.
\newblock URL \url{https://doi.org/10.1016/j.nicl.2017.12.020}.

\bibitem[Pennell(2020)]{pennell}
P.~B. Pennell.
\newblock Unravelling the heterogeneity of epilepsy for optimal individualised
  treatment: advances in 2019.
\newblock \emph{Lancet}, 19, January 2020.
\newblock URL \url{https://doi.org/10.1016/S1474-4422(19)30430-2}.

\bibitem[Team(2018)]{rapids-ai}
RAPIDS~Development Team.
\newblock \emph{RAPIDS: Collection of Libraries for End to End GPU Data
  Science}, 2018.
\newblock URL \url{https://rapids.ai}.

\bibitem[Wang et~al.(2016)Wang, Yan, and Oates]{wang}
Z.~Wang, W.~Yan, and T.~Oates.
\newblock Time series classification from scratch with deep neural networks: A
  strong baseline.
\newblock \emph{arXiv}, December 2016.
\newblock URL \url{https://arxiv.org/abs/1611.06455}.

\bibitem[WHO(2022)]{who}
WHO.
\newblock Epilepsy.
\newblock Fact sheet, World Health Organization, 2022.
\newblock URL \url{https://www.who.int/news-room/fact-sheets/detail/epilepsy}.

\bibitem[Xiao et~al.(2021)Xiao, Li, Wang, Chen, Si, Yao, Li, Duan, and
  Heng]{xiao}
L.~Xiao, C.~Li, Y.~Wang, J.~Chen, W.~Si, C.~Yao, X.~Li, C.~Duan, and P.~Heng.
\newblock Automatic localization of seizure onset zone from high-frequency seeg
  signals: A preliminary study.
\newblock \emph{IEEE Journal of Translational Engineering in Health and
  Medicine}, 9:\penalty0 1--10, 2021.
\newblock URL \url{https://ieeexplore.ieee.org/document/9458292}.

\end{thebibliography}






\end{document}